\documentclass[lettersize,journal]{IEEEtran}
\usepackage{amsmath,amsfonts}
\usepackage{algorithmic}
\usepackage{algorithm}
\usepackage{array}
\usepackage{color,xcolor}
\usepackage[table]{xcolor}
\usepackage{amsmath, amsfonts, amssymb}
\usepackage{algorithm}
\usepackage{dsfont}
\usepackage{mathrsfs}
\usepackage{array}
\usepackage[caption=false,font=normalsize,labelfont=sf,textfont=sf]{subfig}
\usepackage{textcomp}
\usepackage{stfloats}
\usepackage{url}
\usepackage{verbatim}
\usepackage{graphicx}
\usepackage{cite}
\usepackage{booktabs}
\usepackage{multirow}
\usepackage{pifont} 
\usepackage[hidelinks]{hyperref} % 放在最后
\begin{document}

\title{Enhancing Test Time Adaptation with Few-shot Guidance}

\author{%
Siqi Luo,
Yi Xin,
Yuntao Du,
Tao Tan,~\textit{Member, IEEE},\\
~~~~Guangtao Zhai,~\textit{Fellow, IEEE},
and Xiaohong Liu,~\textit{Member, IEEE}%

\thanks{Corresponding author: Xiaohong Liu (e-mail: xiaohongliu@sjtu.edu.cn).}
\thanks{
S. Luo, G. Zhai, and X. Liu are with the Institute of Image Communication and 
Network Engineering, Shanghai Jiao Tong University, Shanghai 200240, China. 
E-mail:\{siqiluo647, zhaiguangtao, xiaohongliu\}@sjtu.edu.cn.
}
\thanks{
Y. Xin is with the State Key Laboratory for Novel Software Technology, 
Nanjing University, Nanjing, China, and with Shanghai Innovation Institute, 
Shanghai, China. 
E-mail:xinyi@smail.nju.edu.cn.
}
\thanks{
Y. Du is with the School of Software \& Joint SDU-NTU Centre for Artificial 
Intelligence Research, Shandong University, Shandong, China. 
E-mail:yuntaodu@sdu.edu.cn.
}
\thanks{
T. Tan is with Macao Polytechnic University, Macao, China. 
E-mail:taotan@mpu.edu.mo.
}
}
% Remember, if you use this you must call \IEEEpubidadjcol in the second
% column for its text to clear the IEEEpubid mark.

\maketitle

\begin{abstract}
Deep neural networks often suffer severe performance degradation when deployed in 
target domains that differ from the training data. Test-Time Adaptation (TTA) seeks to 
alleviate this issue by updating a pre-trained model using only streaming unlabeled target 
data. However, the complete absence of supervision makes it difficult for TTA to correct 
misaligned decision boundaries, often resulting in unstable or even harmful updates under 
distribution shift. We propose Few-Shot Test-Time Adaptation (FS-TTA), a practical extension of TTA that 
leverages a small labeled support set from the target domain prior to deployment. Even 
minimal supervision substantially reduces the uncertainty of blind adaptation and provides 
the boundary information needed for more reliable model updates. Building on this insight, 
we introduce a unified two-stage adaptation framework. Stage I performs fine-tuning for 
boundary alignment using the labeled support samples, further enhanced by feature-level 
diversity to improve robustness in the low-shot regime. Stage II conducts test-time 
distribution refinement using the unlabeled target stream, where reliable pseudo-labels are 
produced via a prototype memory mechanism to ensure stable online adaptation. Extensive experiments on PACS, OfficeHome, and DomainNet demonstrate that the 
proposed FS-TTA setting and framework deliver consistently superior performance and 
significantly improve adaptation reliability over state-of-the-art TTA approaches.
\end{abstract}


\section{Introduction}
\IEEEPARstart{I}{n} recent years, deep neural networks have exhibited remarkable capabilities in representation learning. However, their performance relies heavily on the assumption that the distributions of training (source) and test (target) data are identical~\cite{long2015learning,ganin2015unsupervised,li2017deeper}. In real-world deployment, such a distribution shift is inevitable, as it is practically impossible to collect and annotate data for all possible environments in advance of training. Besides, this distribution shift can significantly degrade the performance of the deployed source model.

To address the aforementioned issues, numerous studies have proposed solutions via domain adaptation~\cite{long2015learning, tzeng2017adversarial, MDD, xiao2021dynamic, xin2023self} and domain generalization~\cite{volpi2018generalizing, zhou2021domain, kim2021selfreg, sicilia2023domain}. While these approaches have demonstrated impressive performance gains on realistic benchmarks, a considerable gap remains between their problem settings and practical application scenarios. Domain adaptation relies on the impractical assumption that target domain data are available and participate in the source training process. 
In contrast, domain generalization aims to directly enhance the generalization of the source model without exploring the target domain data, even if they can be obtained during the test time.

\begin{figure}[t]
    \centering
    \includegraphics[width=1\linewidth]{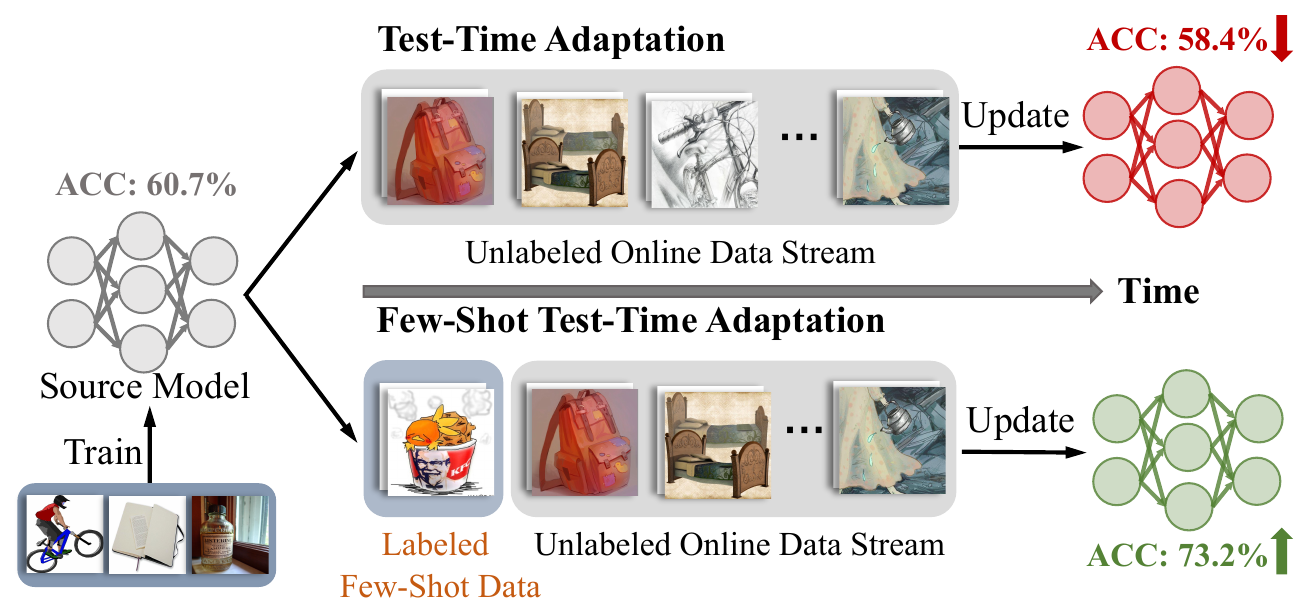}
   \caption{Test Time Adaptation (TTA) vs. Few-Shot Test Time Adaptation (FS-TTA). FS-TTA incorporates a small number of labeled target samples, which can be easily collected offline before deployment with minimal annotation effort, in addition to the unlabeled target data used in TTA. The results for TTA are based on the performance of TENT~\cite{wang2020tent} on the OfficeHome~\cite{venkateswara2017deep}.}
    \label{fig:intro}
    \vspace{-0.1in}
\end{figure}

In order to overcome these limitations of domain adaptation/generalization and protect the privacy of the source data, TENT~\cite{wang2020tent} introduces fully test time adaptation (TTA). TTA aims to adapt a pre-trained source model to the target domain using input mini-batch data during the test time, without relying on source data or supervision. 
TTA is particularly focused on an online setting, where the model must adapt and make predictions immediately upon receiving each batch of potentially non-independent and identically distributed (non-i.i.d.) target samples. To serve this purpose, TENT employs test-time entropy minimization to reduce the generalization error on shifted target data. Additionally, extensive research has sought to improve TTA through various approaches such as pseudo-labeling~\cite{IwasawaM21, wang2023feature}, consistency regularization~\cite{boudiaf2022parameter}, and anti-forgetting regularization~\cite{niu2022efficient}. 
While these methods can perform model adaptation during the test time, they encounter three primary challenges:

~~\textit{1) Domain shift correction:} The certainty of TTA methods in addressing domain shifts effectively without utilizing target labels is questionable. The t-SNE visualization in Figure~\ref{fig:tsne} clearly illustrates this point, where we observe that the feature distribution exhibits negligible change following the adaptation process with TENT. This suggests that TTA methods may struggle to effectively adjust to new domain characteristics in the complete absence of target labels, which could provide essential guidance for adaptation.

~~\textit{2) Generalizability:} The effectiveness of TTA methods varies across different scenarios. In some cases, they might even underperform compared to the pre-trained source model without any adaptation, as illustrated in Figure~\ref{fig:intro} (Source Model vs. TENT). This variability indicates that the generalization performance of TTA methods is not particularly strong and can be influenced by various factors, including the domain shift and the specific characteristics of the model and dataset involved.

~~\textit{3) Data reliance:} The success of TTA methods heavily relies on the availability and quality of unlabeled mini-batch data from the target domain. This reliance presents a challenge, as the adaptation process is directly influenced by the representativeness, quantity, and quality of the available unlabeled data. In scenarios where high-quality, relevant unlabeled data is scarce or not fully representative of the entire target domain, TTA methods may face difficulties in achieving optimal performance, highlighting a major limitation in their application across various real-world settings.

The fundamental difficulty of fully unsupervised test-time adaptation is that the model must 
adjust to distribution shifts while blindly exploring the target domain without any reliable 
supervisory signal. This often results in unstable updates, making such methods difficult to 
deploy in real applications where robustness is essential. In many practical systems, it is 
both feasible and common to obtain a few labeled target samples before deployment, as 
domain experts or users can annotate representative examples during system setup. Even 
such minimal supervision can provide meaningful guidance during the initial stage of 
adaptation. This naturally leads to the question: \textbf{\textit{If given limited supervisory 
information from the target domain, could the adaptation performance be improved?}}

% \begin{figure}[h]
%     \centering
%     \includegraphics[width=1.1\linewidth]{one-shot.pdf}
%     \vspace{-0.8cm}
%     \caption{Performance comparison across different adaptation strategies on OfficeHome. One-shot fine-tuning with a single labeled sample per class already surpasses TENT, showing the effectiveness of minimal supervision.}
%     \label{fig:one-shot}
% \end{figure}

\begin{figure}[!h]
    \centering
    \begin{picture}(0,128)
    \put(-150,-2){
    \includegraphics[width=1.2\linewidth]{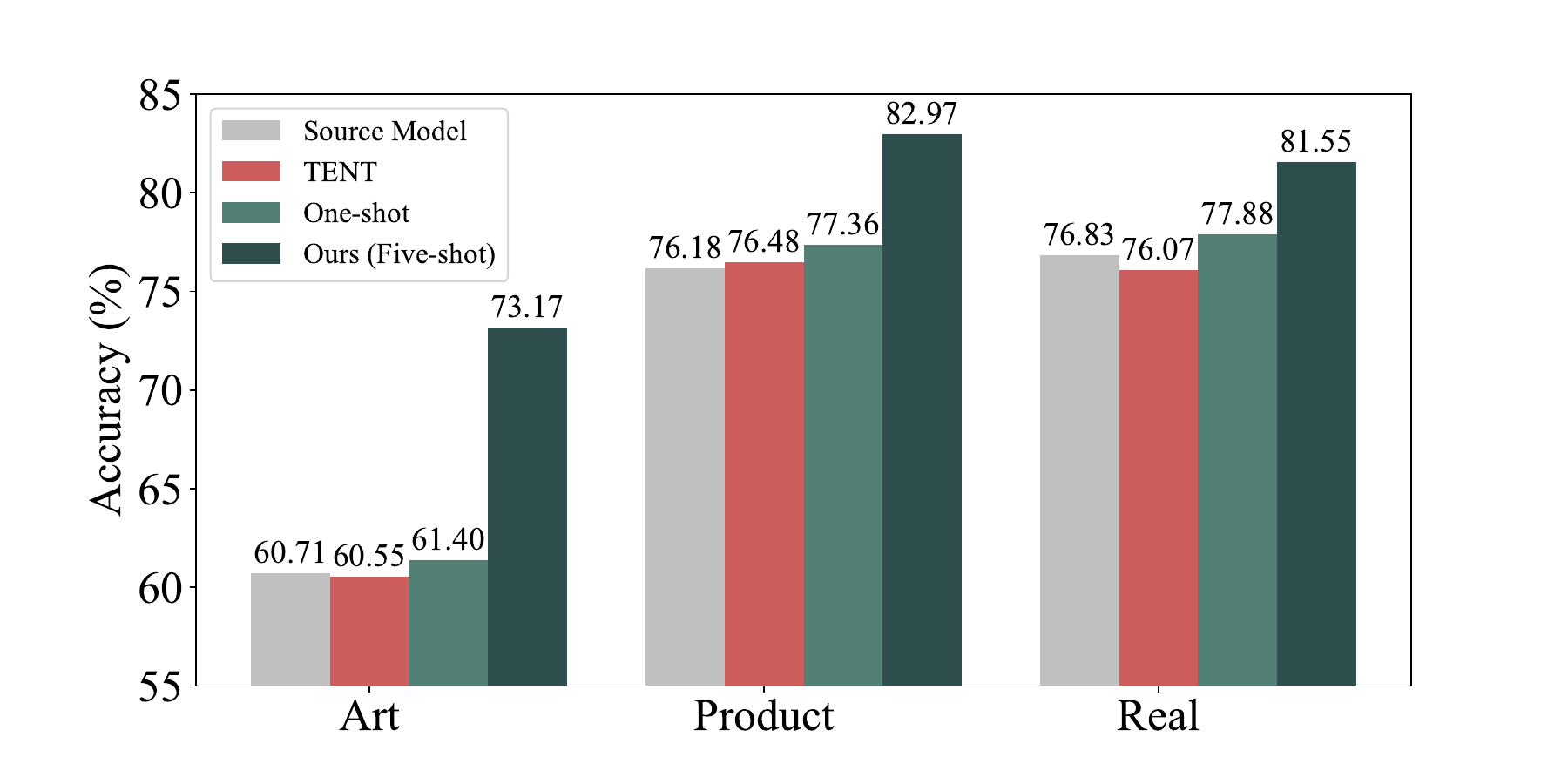}
    }
    \end{picture}
    \vspace{-0.15in}
    \caption{Performance comparison across different adaptation strategies on OfficeHome. One-shot fine-tuning with a single labeled sample per class already surpasses TENT, showing the effectiveness of minimal supervision.}
    \label{fig:one-shot}
\end{figure}

To answer this question, we test the one-shot situation, as shown in Figure~\ref{fig:one-shot}. Specifically, we use one sample per class to fine-tune the source model with cross-entropy loss. We find that the performance is easily improved compared to TENT, which shows that little supervision information can be more effective than a large amount of unsupervised information.

Building on this insight, we propose Few-Shot Test-Time Adaptation (FS-TTA), 
an adaptation framework that fully exploits the complementary roles of the labeled support 
set and the unlabeled test-time stream, encapsulating the principle of \textbf{\textit{few inputs, 
big gains}}. The support set not only alleviates the initial domain shift, but also provides a 
reliable foundation for subsequent adaptation. In \textbf{Stage I: Fine-tuning for Boundary Alignment}, we use the labeled support samples to adjust the 
source model toward the target domain and reinforce this process with a Feature Diversity 
Augmentation (FDA) strategy, which regularizes the low-shot update and yields a more 
reliable initialization. On top of this initialization, \textbf{Stage II: Test-Time Adaptation for Distribution Refinement},  
leverages the unlabeled test-time stream to further adapt the model. The broader 
distributional information contained in the stream complements the limited support set, 
while reliable pseudo-labels, selected according to confidence and consistency criteria, 
ensure that online updates remain stable and resistant to drift under domain shift.

\begin{itemize}
    \item \textbf{Emerging research direction:} We formalize Few-Shot Test-Time Adaptation (FS-TTA), a realistic setting where minimal 
    target supervision is available prior to deployment. This setting bridges the gap between 
    fully unsupervised TTA and traditional adaptation paradigms, highlighting the value of 
    incorporating limited but reliable target-domain signals.
    \vspace{0.15cm}
    \item \textbf{Innovative framework:} We show that FS-TTA naturally decomposes into 
    a unified framework consisting of boundary-level correction and distribution-level 
    refinement, providing a coherent perspective on how labeled and unlabeled target-domain 
    information should be jointly leveraged for stable and reliable adaptation.
    \vspace{0.15cm}
    \item \textbf{State-of-the-art performance:} Extensive empirical evaluations on multiple cross-domain classification benchmarks validate the effectiveness of our framework. Compared to the current state-of-the-art TTA methods, our approach achieves performance gains of 2.0\% on PACS, 7.8\% on OfficeHome, and 3.9\% on DomainNet.   
\end{itemize}

\begin{table*}[t]
\centering
\caption{Comparison with various adaptation settings. where $s$ and $t$ denote source domain and target domain. $L^d$ and $U^d$ denote labeled datasets and unlabeled datasets from domain $d$. ``Online'' means that adaptation can predict a batch of incoming test samples immediately. ``$k$'' represents the number of samples per class. ``C'' indicates the number of classes for the target domain.
}
\renewcommand{\arraystretch}{1.7}
\resizebox{1\textwidth}{!}
{
\begin{tabular}{lccccc}
\toprule
\multirow{2}{*}{\textbf{Setting}}      
& \multirow{2}{*}{\textbf{Source-free}}          
& \multicolumn{3}{c}{\textbf{Training inputs}}      
& \multirow{2}{*}{\textbf{Online}} 
\\ \cmidrule{3-5}
    & & \multicolumn{1}{c}{Source domain(s)}        
    & \multicolumn{1}{c}{Target domain} & \multicolumn{1}{c}{Size of available target data} \\ \midrule
Domain Generalization & \ding{55} & $L^{s_1},\dots,L^{s_N}$  & - & 0 & \ding{55}  \\

Source-Free Domain Adaptation   & \ding{52} & Pre-trained model on $L^{s_1},\dots,L^{s_N}$ & Entire $U^{t}$ & $\|U^{t}\|$ & \ding{55} \\

Few-Shot Transfer Learning   & \ding{52} & Pre-trained model on $L^{s_1},\dots,L^{s_N}$ & Few-shot support set $L^{spt} \subset L^{t}$ & $k \times \text{C}$ & \ding{55} \\

Test Time Adaptation
& \ding{52} & Pre-trained model on $L^{s_1},\dots,L^{s_N}$      & Mini-batch $U^{t}$ & $|\text{mini-batch}|$, typically 128 & \ding{52}

 \\ \midrule
\textbf{Few-Shot Test Time Adaptation}
    & \ding{52} & Pre-trained model on $L^{s_1},\dots,L^{s_N}$ & Few-shot support set $L^{spt} \subset L^{t}$ and mini-batch $U^{t}$ & $k \times \text{C}$ and $|\text{mini-batch}|$ & \ding{52} \\
\bottomrule[1pt]
\end{tabular}
}
\label{tab:settings}
\end{table*}

\section{Related Work}
\label{sec:related work}

\subsection{Domain Generalization}
Domain Generalization (DG) aims to train models on multiple related but distinct source domains to ensure effective performance on unseen target domains. To enhance robustness, DG techniques often employ strategies such as data augmentation~\cite{huang2021fsdr, volpi2018generalizing} and data generation~\cite{zhou2021domain, robey2021model} to introduce greater diversity during training. Other prevalent approaches leverage representation learning to extract domain-invariant features. This includes kernel-based methods~\cite{li2018domain} that project data into a shared feature space, domain adversarial learning~\cite{sicilia2023domain} that aligns distributions via adversarial objectives, and invariant risk minimization~\cite{krueger2021out} which encourages models to perform consistently across domains. In addition, self-supervised~\cite{kim2021selfreg} and meta-learning-based techniques~\cite{chen2022discriminative} have been explored to further improve generalization.  
However, without exposure to the target domain, generalization remains inherently limited.

\subsection{Source-Free Domain Adaptation} 
Source-Free Domain Adaptation (SFDA) aims to adapt a pre-trained source model to an unlabeled target domain while ensuring that no source data is accessed during the adaptation process. By eliminating the dependence on source data, SFDA effectively safeguards source data privacy, making it particularly suitable for scenarios where data sharing is restricted.  SFDA techniques can be broadly categorized into two main approaches: pseudo-label strategies and generative methods. The former leverages target pseudo-labels to facilitate self-training, thereby enabling implicit adaptation without requiring explicit supervision~\cite{tanwisuth2021prototype, ahmed2021unsupervised, liang2021source, xin2023self}. The latter employs generative models to synthesize target-style training data, allowing the model to bridge the domain gap through data augmentation and distribution alignment~\cite{qiu2021source, liu2021source}. Similar to SFDA, our proposed Few-Shot Test Time Adaptation (FS-TTA) also maintains the source-free property, ensuring that adaptation is performed without relying on source data while leveraging a small support set to enhance adaptation efficiency. 

\subsection{Few-Shot Transfer Learning}
Test Time Adaptation (TTA) aims to adapt a pre-trained source model on-the-fly during inference to mitigate distribution shifts. Early TTA methods apply self-supervised learning objectives~\cite{sun2020test}, but typically require access to training data or modification of the training process. TENT~\cite{wang2020tent} addresses this by proposing fully test-time adaptation, relying solely on target samples and adapting batch normalization parameters via entropy minimization. Subsequent approaches such as~\cite{schneider2020improving, nado2020evaluating} update statistics on each incoming mini-batch, while methods like LAME~\cite{boudiaf2022parameter} and EATA~\cite{niu2022efficient} tackle catastrophic forgetting during continual adaptation. TSD~\cite{wang2023feature} further integrates self-training to selectively update using confident predictions.  
As a result, adaptation often relies heavily on the quality of incoming test samples.

\subsection{Test Time Adaptation}
Test Time Adaptation (TTA) aims to adapt a pre-trained source model during inference to mitigate distribution shifts between training and test domains. Early TTA methods address this challenge through self-supervised auxiliary tasks~\cite{sun2020test}, which, while effective, often require access to training data or modifications to the training procedure. To overcome this limitation, TENT~\cite{wang2020tent} proposes fully test-time adaptation by leveraging only target data, updating batch normalization parameters via entropy minimization. Building on this, subsequent works~\cite{schneider2020improving, nado2020evaluating} estimate batch normalization statistics dynamically from incoming test batches. Other approaches, such as LAME~\cite{boudiaf2022parameter} and EATA~\cite{niu2022efficient}, focus on preventing catastrophic forgetting during continuous model updates. More recently, TSD~\cite{wang2023feature} incorporates self-training by selecting confident test samples to guide adaptation.  
Despite their progress, these methods heavily rely on the quality and stability of online target data.

\subsection{Comparisons with Other Settings}
We compare \textit{Few-Shot Test Time Adaptation} (FS-TTA) with similar problem settings (details are in the appendix), as illustrated in Table~\ref{tab:settings}.
\begin{itemize}
    \item Compared with \textit{Domain Generalization}, FS-TTA eliminates the necessity of accessing source data, thereby ensuring the preservation of source data privacy. Moreover, it allows for adaptation to the downstream target domain by updating model parameters, making it more flexible and applicable in real-world settings. 
    \vspace{0.1cm}
    \item Compared with \textit{Source-Free Domain Adaptation}, FS-TTA removes the constraint of requiring all target domain data to be available at once. Instead, it facilitates dynamic and continuous online model updates, enabling adaptation based on incoming mini-batches of target data, which is particularly beneficial in streaming or real-time applications.
    \vspace{0.1cm}
    \item Compared with \textit{Few-Shot Transfer Learning}, FS-TTA not only makes use of a limited number of target domain samples for adaptation but also continuously refines the model during test time by incorporating online mini-batch target data. This ensures more efficient and progressive adaptation to changing target distributions.
    \vspace{0.1cm}
    \item Compared with \textit{Test Time Adaptation}, FS-TTA leverages a small auxiliary set of target samples, allowing the pre-trained source model to adapt more quickly and effectively to the target domain. Additionally, FS-TTA demonstrates superior performance in handling challenging scenarios where there are substantial domain shifts, making it a more robust and reliable solution.
\end{itemize}

\begin{figure*}[t]
    \centering
    \includegraphics[width=1 \linewidth]{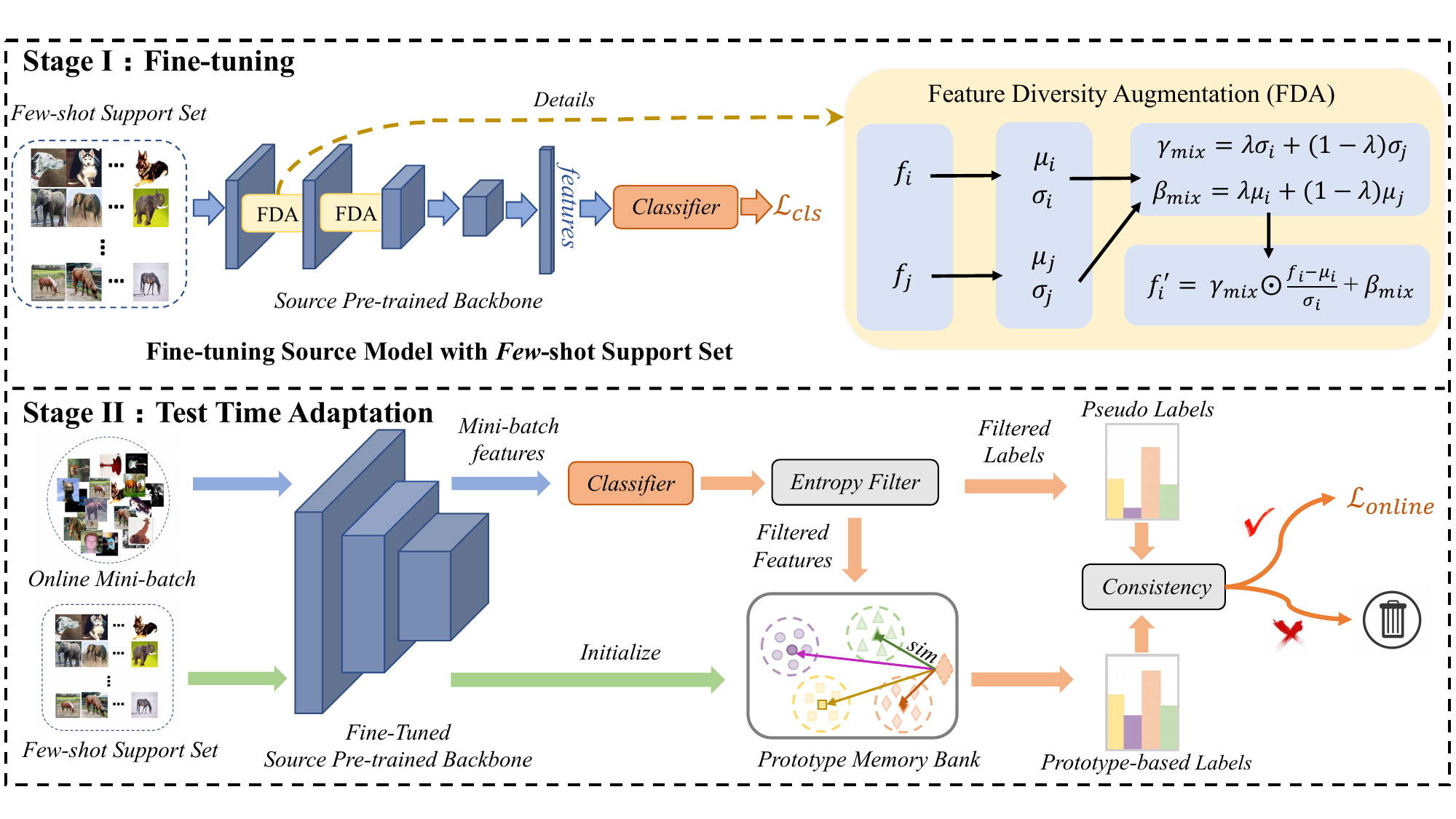}
    \vspace{-0.8cm}
    \caption{Illustration of our two-stage framework. In Stage I, we employ the few-shot support set to fine-tune the source model. To prevent overfitting, we propose FDA module. In Stage II, we maintain a prototype memory bank to guide test time adaptation. In order to update the prototype memory bank and model with effective samples, we propose the entropy filter and consistency selection modules.
    }
    \label{fig:approach}
\end{figure*}
\section{Preliminary}
\subsection{Instance Normalization}
Instance Normalization~\cite{ulyanov2016instance} is a normalization technique widely used in deep neural network architectures, especially in the context of style transfer and generative models.  Let us consider a batch images with size $N \times C \times H \times W$, where $N$ is the batch size, $C$ is the number of channels, and $H$ and $W$ are the height and width of the images. For each sample $i$ and channel $c$, we compute the mean and standard deviation as follows:
\begin{equation}
 \mu_{i,c} = \frac{1}{H \times W} \sum_{h=1}^H \sum_{w=1}^W x_{i,c,h,w},
 \label{eq:mu}
\end{equation}
\begin{equation}
 \sigma_{i,c} = \sqrt{\frac{1}{H \times W} \sum_{h=1}^H \sum_{w=1}^W (x_{i,c,h,w} - \mu_{i,c})^2},
 \label{eq:sig}
\end{equation}
where $x_{i,c,h,w}$ denotes the input feature of samples $i$, channel $c$, height $h$, and width $w$. After computing the mean and standard deviation, we can normalize the input features:
\begin{equation}
{\rm IN}(x_{i,c,h,w}) = \gamma \frac{x_{i,c,h,w} - \mu_{i,c}}{\sigma_{i,c}} + \beta ,
\end{equation}
where $\gamma, \beta \in R^{C}$ are learnable transformation parameters.

\subsection{Class Prototype}
The class prototype is a representative point in the feature space that summarizes the key characteristics of a class. For each class, it serves as a centroid or an anchor point around which samples of the class cluster. Let us denote $F = \{f_1, f_2, ..., f_n\}$ as a set of $n$ sample features in class $c$, where each $f_i \in R^d$ represents a $d$-dimensional feature vector of a sample. The prototype $P$ of the class $c$ is calculated as the mean of all feature vectors, namely that:
\begin{equation}
    P_c = \frac{1}{n} \sum_{i=1}^{n} f_i,
    \label{equ:class_prototype}   
\end{equation}
where $P_c \in R^d$. The class prototype plays an important role in few-shot scenarios.

\section{Method}
\label{sec:method}
\subsection{Problem Setting}
Considering a typical scenario where a source model $f_{\theta_{s}}$ is equipped with parameters $\theta_{s}$ and trained on source datasets $\mathcal{D}{s_1}, \mathcal{D}{s_2}, …, \mathcal{D}{s_n}$, our objective is to adapt this pre-trained model to a target domain $D_t$ without accessing source data. A small, labeled support set $S = {(s_i, y_i)}$ is provided from $D_t$, where $s_i$ denotes an image and $y_i$ its corresponding label. During test time, unlabeled target samples arrive sequentially in mini-batches. Few-Shot Test Time Adaptation (FS-TTA) aims to effectively adapt the source model $f_{\theta_{s}}$, by leveraging the support set $S$ in conjunction with the streaming unlabeled data to mitigate domain shift. Notably, the support set $S$ can be acquired offline prior to deployment, and in many real-world applications, collecting such limited supervision is both feasible and cost-efficient.

\subsection{Stage I: Boundary Alignment via Few-Shot Fine-Tuning} 
A major challenge in test-time adaptation is that the source model’s decision boundary does 
not align with the target-domain distribution. The few-shot support set provides the only 
unbiased supervision capable of correcting this boundary misalignment.Therefore, Stage I aims to align the model 
boundary with the target domain by fine-tuning the pre-trained source model on these 
labeled samples.
 Given the limited number of samples per class, there is a potential risk of overfitting during the fine-tuning process. To mitigate this, we introduce the Feature Diversity Augmentation (FDA) module, which generates new features by mixing statistics. Ultimately, we use a supervised classification loss to fine-tune the pre-trained source model. This entire procedure is illustrated in Stage I of Figure~\ref{fig:approach}.

 \vspace{0.1cm} \noindent \textbf{Feature Diversity Augmentation (FDA).}
Prior research~\cite{zhou2021domain} has demonstrated a significant association between feature statistics and image style, which is intricately linked to data distribution within the field of computer vision. To increase style diversity while preserving semantic consistency, we introduce Feature Diversity Augmentation (FDA), a feature-level data augmentation technique that simulates various image styles without altering the original class labels. This approach effectively enriches the support set and helps reduce the risk of overfitting during fine-tuning.

FDA is incorporated between layers (blocks) in the pre-trained source backbone, as depicted in Figure~\ref{fig:approach}. More specifically, FDA mixes the feature statistics of two random samples to generate new features. The computations within the FDA module can be summarized in three steps. Firstly, given two feature maps $f_i$ and $f_j$ from the support set, we compute their feature statistics ($\mu_i,\sigma_i$) and ($\mu_j,\sigma_j$). Secondly, FDA generates the mixtures of feature statistics:

\begin{equation}
\boldsymbol{\gamma}_{\text {mix }}  =\lambda \sigma _i+(1-\lambda) \sigma _j,
\end{equation}
\begin{equation}
    \boldsymbol{\beta}_{\text {mix }}  =\lambda \mu_i+(1-\lambda) \mu_j.
\end{equation}

In this case, $\lambda$ denotes the mixing ratio coefficient. Ultimately, the mixtures of feature statistics are applied to the feature map $f_i$ via instance normalization:
\begin{equation}
f_i^{\prime}=\boldsymbol{\gamma}_{\operatorname{mix}} \odot \frac{f_i-\mu_i}{\sigma_i}+\boldsymbol{\beta}_{\operatorname{mix}},
\end{equation}
where $f_i^{\prime}$ represents the newly generated feature map.

 \vspace{0.1cm} \noindent \textbf{Fine-Tuning Source Model.}
To enhance the adaptation of the pre-trained source model to the target, we employ the few-shot support set to fine-tune the model with the FDA module. Specifically, the few-shot support set is processed through $f_{\theta_{s}}$ to minimize a supervised loss, defined as:
\begin{equation}
    \mathcal{L}_{\text {cls}}=-\sum_{i=1}^{k*C}\mathcal{H}\left(y_{i}, p\left(\hat{y_i} \mid \boldsymbol{s}_{i}\right)\right),
\label{equ:cross-entropy}
\end{equation}
where $\mathcal{H}(\cdot)$ is the cross-entropy loss. The term $y_i$ is the ground-truth label of $s_i$, indicating one of sample from few-shot support set, and $C$ represents categories of the target.

\subsection{Stage II: Test Time Adaptation}
After obtaining a target-aware initialization from Stage I, the model encounters a stream of 
unlabeled target samples $x=\{x_1, x_2, \ldots, x_B\}$ during deployment. These samples provide additional distributional insights beyond what the support set can offer. Stage II aims to refine the model online by leveraging this unlabeled stream while 
ensuring stable updates under distribution shift.

The central concept of this stage is to employ a self-training strategy to update the fine-tuned source model online, enabling it to fully adapt to the target domain. This involves assigning pseudo-labels to unlabeled online mini-batches and using these labels to further update the model. Thus, we first generate the pseudo-labels by $\hat{y_i}=\text{argmax}(p_i)$ for $x_i$, where $p_i$ is the prediction logits.
% produced by the fine-tuned source model. 
However, it is inevitable that there are always some noisy samples are misclassified, leading to wrong pseudo-labels. To address this issue, we propose two modules to produce high quality pseudo-labels. The first is entropy filter, which screens out unreliable samples using Shannon entropy~\cite{shannon2001mathematical}. Typically, samples with higher entropy are considered to have lower prediction confidence. The second module is a prototype memory bank classification, which works in tandem with the classifier. The prototype memory bank is used to generate pseudo-labels outside the classifier, according to the nearest class prototype in the feature space. After that, pseudo-labels with consistency prediction is preserved for model adaptation.  The entire process is outlined in Stage II of Figure~\ref{fig:approach}.

 \vspace{0.1cm} \noindent \textbf{Entropy Filter.} To dynamically update the model using online mini-batch target, it is crucial to filter out noisy samples, as they may be assigned to incorrect classes, resulting in inaccurate prototype computation. In this regard, we propose the Entropy Filter, which employs Shannon entropy~\cite{shannon2001mathematical} to select confident samples in the mini-batch. For an sample $x_{i}$, its entropy can be computed as:
\vspace{-0.1cm}
\begin{equation}
    H(p_{i}) = -\sum (p_{i})\cdot log(p_{i}).
\end{equation}

Based on the insights from previous work~\cite{wang2020tent}, high entropy samples should be filtered out, as lower entropy typically indicates higher accuracy. Consequently, we sort the entropy of all samples in the mini-batch and select the top $\alpha\%$ samples with lower entropy, donated as $\hat{x} = \{\hat{x}_1, \hat{x}_2, \ldots, \hat{x}_{\lfloor \alpha \cdot B \rfloor}\}$.

 \vspace{0.1cm} \noindent \textbf{Prototype Memory Bank.} We maintain a prototype memory bank $M=\{m_1, m_2, ..., m_C\}$ to store class prototypes, where $C$ represents categories of the target. The prototype memory bank is initialized with the few-shot support set $S$, defined as:

\vspace{-0.1cm}
\begin{equation}
    m_{c_0} = \frac{\sum_{i=1}^{\lvert S \rvert} f_i \cdot \mathds{1}[y_i = c]}{\sum_{i=1}^{\lvert S \rvert} \mathds{1}[y_i = c]},
    \label{eq:proto}
\end{equation}
where $\mathds{1}[\cdot]$ represents an indicator function, yielding a value of 1 if the argument is true or 0 otherwise, and $m_{c_0}$ denotes the initial moment of the $c$-th class prototype. Thanks to the few-shot support, precise guidance can be provided during the initial phase, thereby reducing reliance on the quality of online mini-batch data.

Throughout the test time adaptation process, we persistently update the prototype memory bank by incorporating selected reliable samples with pseudo labels:
\vspace{-0.2cm}
\begin{equation}
    m_{c_t} = \beta \cdot m_{c_{t-1}} + (1 - \beta) \cdot 
    \frac{\sum_{j=1}^{\lvert \hat{x} \rvert } f_j \cdot \mathds{1}[\hat{y}_j = c]}{\sum_{j=1}^{\lvert \hat{x} \rvert} \mathds{1}[\hat{y}_j = c]}.
\end{equation}
where $m_{c_t}$ represents the $c$-th class prototype at time $t$, and $\beta$ represents the sliding update coefficient.

 \vspace{0.1cm} \noindent \textbf{Test Time Adaptation.}
During the test time adaptation, we adopt high-quality pseudo-labeled samples to guide the model update. First, we define the prototype-based classification output as the softmax over the feature similarity to prototypes for class $c$:
\begin{equation}
    \hat{p}_{j}^{c}=\frac{\exp \left(\operatorname{sim}\left(f_{j}, m_{c}\right)\right)}{\sum_{c=1}^{C} \exp \left(\operatorname{sim}\left(f_{j}, m_c\right)\right)},
\end{equation}
where $sim(\cdot,\cdot)$ represents cosine similarity. Subsequently, we propose that, for a reliable sample, the outputs of the fine-tuned model and prototype-based classification should be similar. Therefore, we propose the consistency filter to identify incorrect predictions. This strategy can be implemented through a filter mask for samples $x_j$ as follows:
\begin{equation}
\mathcal{M}_{j}=\mathds{1}[\arg \max p_{j}=\arg \max \hat{p}_{j}].
\end{equation}

Ultimately, we can update the model using reliable samples, and the loss can be formulated as follows:
\begin{equation}
    \mathcal{L}_{online}=\frac{ {\textstyle \sum_{j=1}^{\| \hat{x}\|}} \mathcal{H}_{j} * \mathcal{M}_{j} }{{\textstyle \sum_{j=1}^{\| \hat{x}\|}}\mathcal{M}_{j}}.
\end{equation}
% \begin{equation}
%     \mathcal{L}_{\text{online}} = \frac{\displaystyle\sum_{j=1}^{|\hat{x}|} \mathcal{H}_{j} \cdot \mathcal{M}_{j}}{\displaystyle\sum_{j=1}^{|\hat{x}|} \mathcal{M}_{j}}.
% \end{equation}

It's noteworthy that our self-training process does not involve specifying any threshold, which enhances the model's generalizability.

\begin{table*}[!t]
\centering
% \vspace{-1cm}
\caption{Comparison with test-time adaptation methods on three datasets with ResNet-50 backbone. FS-TTA achieves consistent improvements over TSD~\cite{wang2020tent}, the strongest baseline method.}
\resizebox{0.95\textwidth}{!}{
% \setlength
% \tabcolsep{1pt} % default value: 6pt
\setlength
\tabcolsep{3pt} % default value: 6pt
\begin{tabular}{*{12}{lccccccccccc}}
\toprule

%%%%%%%%%%%%%%%%%%%%%%%%%
\multicolumn{1}{c}{}  
& \multicolumn{5}{c}{\textbf{OfficeHome}}
& \multicolumn{5}{c}{\textbf{PACS}}
& \multicolumn{1}{c}{\textbf{DomainNet}}
\\ 
\cmidrule(lr){1-1}
\cmidrule(lr){2-6} 
\cmidrule(lr){7-11}
\cmidrule(lr){12-12}
%%%%%%%%%%%%%%%%%%%%%%%%%
\textbf{Method}
& \multicolumn{1}{c}{\textbf{Art}} 
& \multicolumn{1}{c}{\textbf{Clip}}
& \multicolumn{1}{c}{\textbf{Prod}} 
& \multicolumn{1}{c}{\textbf{Real}}
& \multicolumn{1}{c}{\textbf{Avg.}}

& \multicolumn{1}{c}{\textbf{Art}} 
& \multicolumn{1}{c}{\textbf{Cart}} 
& \multicolumn{1}{c}{\textbf{Phot}}
& \multicolumn{1}{c}{\textbf{Sket}} 
& \multicolumn{1}{c}{\textbf{Avg.}} 

& \multicolumn{1}{c}{\textbf{Avg.}}
\\ 
\cmidrule(lr){1-1}
\cmidrule(lr){2-6} 
\cmidrule(lr){7-11}
\cmidrule(lr){12-12}
%%%%%%%%%%%%%%%%%%%%%%%%%
\multicolumn{12}{l}{\textit{Test time adaptation methods}}\\
ERM~\cite{vapnik1999overview} & 60.7 & 55.7 & 76.2 & 76.8 &67.4 &82.5 &80.8 &94.0 &80.9 &84.5 &45.2\\

BN~\cite{nado2020evaluating} & 58.2 & 55.6 & 75.1 & 75.5 &66.1 &83.2 &84.9 &94.0 &77.9 &85.0 &43.3\\

TENT~\cite{wang2020tent} & 60.6 & 58.7 & 76.5 & 76.1 &68.0 &85.2 &86.7 &94.9 &82.9 &87.4 &44.7\\

T3A~\cite{IwasawaM21} & 61.2 & 56.7 & 78.0 & 77.3 &68.3 
&84.0 &82.3 &95.0 &82.7 & 86.0 &46.1\\

ETA~\cite{niu2022efficient} & 58.4 & 55.8 & 75.2 & 75.5 &66.2 &83.2 &84.9 &94.0 &77.9 &85.0 &46.1\\

LAME~\cite{boudiaf2022parameter} & 58.7 & 55.6 & 75.1 & 75.4 &66.2 &84.9 &85.5 &95.0 &80.9 &86.6 &43.2\\ 

PROGRAM~\cite{sun2024program} & 63.4 &  54.3 &  77.2 &  77.2  & 68.0 &87.2 & 84.1 &96.9 & 76.4 &86.2 &43.3\\

DEYO~\cite{lee2024entropy} & 63.8 &  54.9 &  76.4 &  77.3  & 68.1 &88.4 & 85.2 &\underline{97.1} & 82.3 &88.2 &42.5 \\

TSD~\cite{wang2023feature} & 62.3 & 57.5 & 77.5 & 77.5  & 68.7 &87.6 &\underline{88.7} &96.1 &85.0 &89.4 &47.7\\

\cmidrule(lr){1-1}
\cmidrule(lr){2-6} 
\cmidrule(lr){7-11}
\cmidrule(lr){12-12}
\multicolumn{12}{l}{\textit{Fine-tuning + Test time adaptation methods}}\\

FT+TENT~\cite{wang2020tent}               
& 68.8
& \underline{65.5}
& 79.8             
& 78.5
& 73.2
& 87.0          
& 86.9
& 95.2
& 83.6
& 88.2
& 45.4
\\
FT+TSD~\cite{wang2023feature}              
& \underline{70.5}
& 65.1
& \underline{80.3}            
& \underline{79.2}
& \underline{73.8}
& \underline{88.3}         
& 88.6
& 96.5
& \underline{85.9}
& \underline{89.8}
& \underline{48.5}
\\

\cmidrule(lr){1-1}
\cmidrule(lr){2-6} 
\cmidrule(lr){7-11}
\cmidrule(lr){12-12}

FS-TTA                   
& \textbf{73.2}
& \textbf{68.3}
& \textbf{83.0}             
& \textbf{81.6}
& \textbf{76.5}
& \textbf{90.4}          
& \textbf{89.7} 
& \textbf{97.6}
& \textbf{87.8}
& \textbf{91.4}
& \textbf{51.6}
\\
\rowcolor[gray]{0.95}
$\Delta_{up}$ over TSD         
& \textcolor{red}{(+10.9)$\uparrow$}
& \textcolor{red}{(+10.8)$\uparrow$}
& \textcolor{red}{(+5.5)$\uparrow$}            
& \textcolor{red}{(+4.1)$\uparrow$} 
& \textcolor{red}{(+7.8)$\uparrow$} 
& \textcolor{red}{(+2.8)$\uparrow$}       
& \textcolor{red}{(+1.0)$\uparrow$}
& \textcolor{red}{(+1.5)$\uparrow$}
& \textcolor{red}{(+2.8)$\uparrow$}
& \textcolor{red}{(+2.0)$\uparrow$}
& \textcolor{red}{(+3.9)$\uparrow$}
\\
\bottomrule
\end{tabular}
}
\label{tab:compare_tta}
\vspace{-0.1in}
\end{table*}

\section{Experiment}
\label{sec:experiment}
\subsection{Experimental Settings}
 \vspace{0.1cm} \noindent \textbf{Dataset.} To evaluate the effectiveness of our proposed setting and method, we conduct experiments on three cross-domain benchmarks. 
\begin{itemize}
    \item \textbf{PACS}~\cite{li2017deeper} consists of 9,991 images spanning four distinct domains: Art, Cartoon (Cart), Photo (Phot), and Sketch (Sket). Each domain contains seven object categories: dog, elephant, giraffe, guitar, horse, house, and person.
    \item \textbf{Office-Home}~\cite{venkateswara2017deep} comprises 15,588 images distributed across four domains: Art, Clipart (Clip), Product (Prod), and Real-World (Real), with each domain encompassing 65 image categories.
    \item \textbf{DomainNet}~\cite{peng2019moment} is a large-scale dataset containing six domains: Clipart (Clip), Infograph (Info), Painting (Pain), Quickdraw (Quic), Real, Sketch (Sket), comprising a total of 586,575 images across 345 classes.
\end{itemize}
  
\vspace{0.1cm} \noindent \textbf{Implementation Details.} In our main experiments, we employ ResNet-50\cite{he2016deep}, pre-trained on ImageNet-1k\cite{russakovsky2015imagenet}, as the backbone model, as it is widely adopted in the test-time adaptation literature. For source model training, we follow the leave-one-domain-out protocol, as recommended by prior studies~\cite{wang2023feature,zhou2021domain}, treating one domain as the unlabeled target and the rest as source domains.We set the batch size to 32 for each source domain and use a learning rate of 5e-5. Both the dropout probability and weight decay are set to zero. The source model is trained for 5,000 iterations, except for DomainNet, where we extend training to 15,000 iterations, following the methodology in~\cite{swad}. All images are resized to 224 × 224, and data augmentation is applied during source domain training, including random cropping, horizontal flipping, color jittering, and intensity adjustments. For few-shot test time adaptation, we also employ the Adam optimizer and set the batch size to The few-shot support set typically selects 5 to 16 samples per class, depending on the difficulty of the target. We carry out all experiments on NVIDIA V100 GPUs.

% For the implementation of different few-shot transfer learning methods, we utilized the publicly released code of LCCS\footnote{https://github.com/zwenyu/lccs}~\cite{zhang2022few}.

% For source model training, we adhere to the leave-one-domain-out protocol, as recommended by previous work ~\cite{wang2023feature,zhou2021domain}. We employ the Adam optimizer with a learning rate of $5e^{-5}$. For few-shot test time adaptation, we also employ the Adam optimizer and set the batch size to The few-shot support set typically selects 5 to 16 samples per class, depending on the difficulty of the target. We carry out all experiments on NVIDIA V100 GPUs.

\begin{figure}[t]
    \centering
    \vspace{-0.2in}
    \includegraphics[width=0.82\linewidth]{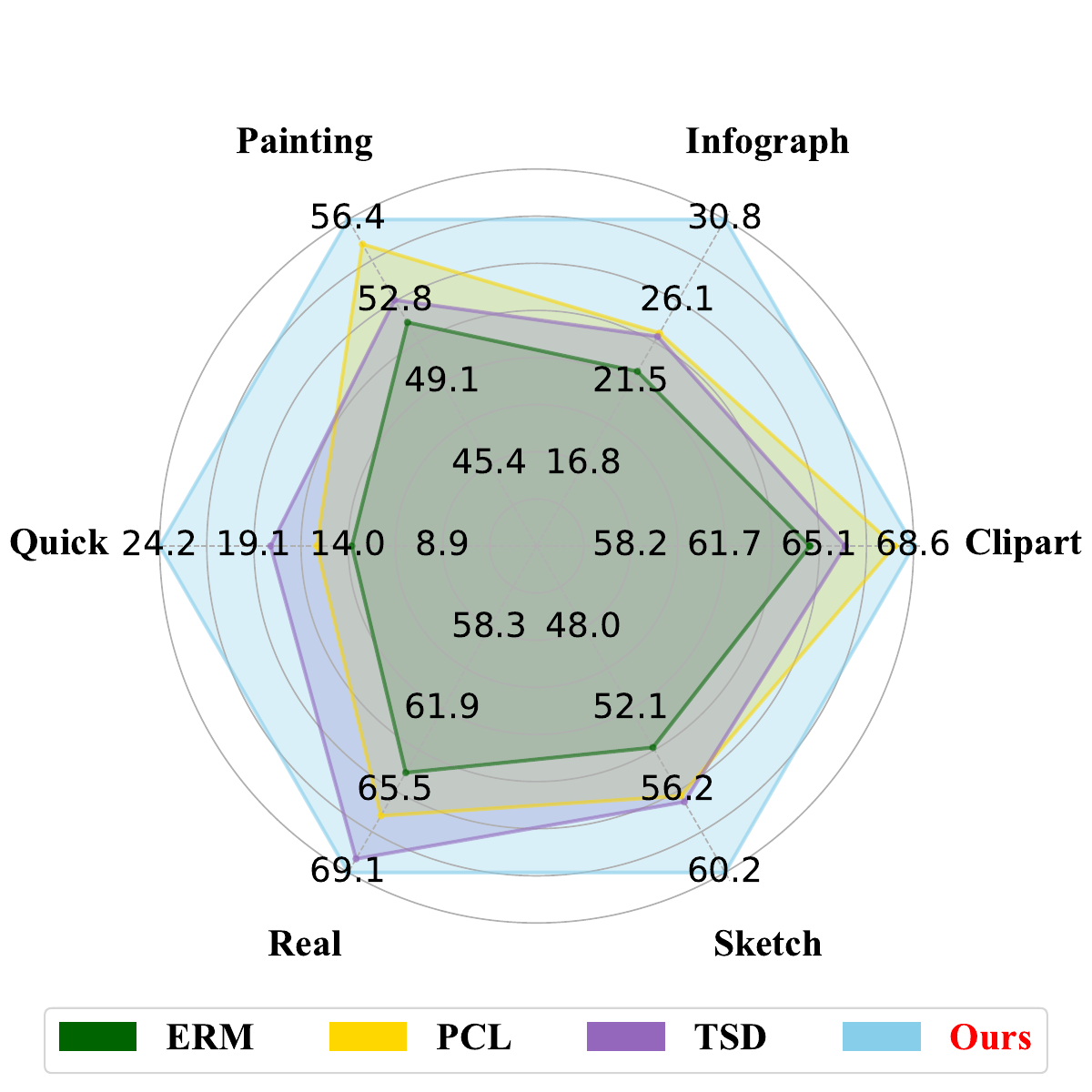}
    \vspace{-0.05in}
    \caption{Comprehensive comparison between our method and the state-of-the-art method in DG/TTA settings on DomainNet.}
    \label{fig:exp_performance}
    \vspace{-0.15in}
\end{figure}

\begin{table*}[t]
\centering
\caption{Compared with existing DG and SFDA methods on OfficeHome and DomainNet.
}
\resizebox{0.95\textwidth}{!}{
\setlength\tabcolsep{6pt} % default value: 6pt
\begin{tabular}{lcccccccccccc}
\toprule
  
& \multicolumn{5}{c}{\textbf{OfficeHome}}
& \multicolumn{7}{c}{\textbf{DomainNet}} \\ 
\cmidrule(lr){2-6}
\cmidrule(lr){7-13}

\textbf{Method} 
& \textbf{Art} 
& \textbf{Clip}
& \textbf{Prod} 
& \textbf{Real}
& \textbf{Avg.}
& \textbf{Clip} 
& \textbf{Info} 
& \textbf{Pain} 
& \textbf{Quic}
& \textbf{Real}
& \textbf{Sket}
& \textbf{Avg.}
\\ 
\midrule
\multicolumn{6}{l}{\textit{Domain generalization methods}}\\
ERM~\cite{vapnik1999overview} 
& 60.7  & 55.7  & 76.2  & 76.8 & 67.4 
& 64.8 & 22.1 & 51.8 & 13.8 & 64.7 & 54.0 & 45.2\\

DNA~\cite{chu2022dna}
& 67.7  & 57.7  & 78.9  & 80.5  & 71.2 
& 66.1  & 23.0 & 54.6  & 16.7  & 65.8  & 56.8  & 47.2\\

PCL~\cite{yao2022pcl}  
& 67.3   & 59.9  & 78.7 & 80.7  & 71.6 
& 67.9  & 24.3  & 55.3  & 15.7  & 66.6  & 56.4  & 47.7\\

SWAD~\cite{swad} 
& 66.1  & 57.7  & 78.4  & 80.2  & 70.6 
& 66.1  & 22.4  & 53.6  & 16.3  & 65.5  & 56.2  & 46.7\\ 
\midrule
\multicolumn{6}{l}{\textit{Source-free domain adaptation methods}} \\
F-mix~\cite{KunduKBMKJR22}
& 72.6 & 67.4 & \underline{85.9} & \underline{83.6} & \underline{77.4} 
& \textbf{75.4} & 24.6  & \underline{57.8}  & 23.6  & 65.8  & 58.5  & 51.0\\
\midrule
FS-TTA              
& \underline{73.2}
& \underline{68.3}
& 83.0    
& 81.6
& 76.5
& 68.6               
& \underline{30.8}
& 56.4
& \underline{24.2}
& \underline{69.1}
& \underline{60.2}     
& \underline{51.6}
\\
SWAD + FS-TTA             
& \textbf{77.4}
& \textbf{71.1}
& \textbf{86.4}     
& \textbf{84.2}
& \textbf{79.8}
& \underline{69.2}
& \textbf{31.5}
& \textbf{57.9}
& \textbf{25.1}
& \textbf{70.8}
& \textbf{62.0}
& \textbf{52.8}
\\
\bottomrule
\end{tabular}
}
\label{tab: compare_dg}
\vspace{-0.2in}
\end{table*}
 \vspace{0.1cm} \noindent \textbf{Baselines.} We compare our method with various test-time adaptation (TTA) approaches, including BN\cite{nado2020evaluating}, TENT\cite{wang2020tent}, ETA\cite{niu2022efficient}, T3A\cite{IwasawaM21}, LAME\cite{boudiaf2022parameter}, TSD\cite{wang2023feature},PROGRAM~\cite{sun2024program} and DEYO~\cite{lee2024entropy}. Additionally, we establish new baselines by integrating fine-tuning with existing TTA methods to ensure a more comprehensive comparison.
Furthermore, we compare our approach with selected methods from domain generalization, source-free domain adaptation, including DNA\cite{chu2022dna}, PCL\cite{yao2022pcl}, SWAD\cite{swad}, and F-mix\cite{KunduKBMKJR22}. Finally, we set up a comparison with the few-shot transfer learning methods, including AdaBN~\cite{li2016AdaBN}, $L^2$~\cite{li2018explicitTL}, DELTA~\cite{Li2019DELTADL}, FLUTE~\cite{triantafillou2021flute}, LCCS~\cite{zhang2022few}. For a global overview,  we compare our method with state-of-the-art method in various settings, as shown in Figure~\ref{fig:exp_performance}.

\vspace{-0.2cm}
\subsection{Performance Comparisons}
 \vspace{0.1cm} \noindent \textbf{Comparison with TTA methods.}
Table~\ref{tab:compare_tta} details the comparison results between our method and various TTA methods on the Office-Home and PACS datasets, as well as the final results of DomainNet (detailed in Table~\ref{tab: compare_dg}). We observe that our method achieves state-of-the-art performance. 

Primarily, our approach exhibits a significant enhancement in performance compared to the source model (ERM). Our FS-TTA achieves improvements across all four tasks on Office-Home, with gains of 12.5\% (Art), 12.6\% (Clipart), 6.8\% (Product), and 4.8\% (Real), respectively. Notably, our method demonstrates more substantial improvement on the more challenging tasks (\textit{e.g.}, Art and Clipart), confirming that FS-TTA is more friendly for large domain shifts. On the other two datasets, we observe average performance increments of 6.9\% (PACS) and 6.4\% (DomainNet). 

\begin{table}[t]
% \vspace{-1.25cm}
\caption{Compared with few-shot transfer learning methods on PACS dataset.}
\vspace{-0.1cm}
\centering
\resizebox{0.48\textwidth}{!}{
\setlength
\tabcolsep{9pt} % default value: 6pt
\begin{tabular}{*{6}{lccccc}}
\toprule
%%%%%%%%%%%%%%%%%%%%%%%%%
  
\multicolumn{1}{c}{} 
& \multicolumn{5}{c}{\textbf{PACS}}     
\\ 
\cmidrule(lr){1-1}
\cmidrule(lr){2-6} 
%%%%%%%%%%%%%%%%%%%%%%%%%

\textbf{Method} 
& \multicolumn{1}{c}{\textbf{Art}} 
& \multicolumn{1}{c}{\textbf{Cart}}
& \multicolumn{1}{c}{\textbf{Phot}} 
& \multicolumn{1}{c}{\textbf{Sket}}
& \multicolumn{1}{c}{\textbf{Avg.}}
\\ 
\cmidrule(lr){1-1}
\cmidrule(lr){2-6}
%%%%%%%%%%%%%%%%%%%%%%%%%
\multicolumn{6}{l}{\textit{Few-shot transfer learning methods}}\\
AdaBN~\cite{li2016AdaBN} &85.0   &83.5   &96.0   &78.7  &85.8 \\

$L^2$~\cite{li2018explicitTL} &85.6   &84.1   &96.4   &76.3  &85.6\\

DELTA~\cite{Li2019DELTADL} &85.6   &83.8   &96.5  &76.3  &85.6 \\

FLUTE~\cite{triantafillou2021flute} &87.2   &86.1   &97.2   &81.7  &88.1 \\

LCCS~\cite{zhang2022few} &\underline{87.7}   &\underline{86.9}   &\underline{97.5}   &\underline{83.0}  &\underline{88.8} \\

\midrule

FS-TTA             
& \textbf{90.4}
& \textbf{89.8}
& \textbf{97.6}     
& \textbf{87.9}
& \textbf{91.4}
\\

\bottomrule
\end{tabular}
}
\label{tab:compare_fewshot}
\vspace{-0.15in}
\end{table}

Moreover, our method outperforms the state-of-the-art TTA method, TSD, with average performance increments of 2.0\% (PACS), 7.8\% (Office-Home), and 3.9\% (DomainNet). The lesser improvement in PACS can be attributed to its lower complexity, while our method shows superior performance on the more challenging Office-Home and DomainNet datasets. This significant improvement benefits from our effective utilization of few-shot target information, including the FDA module and initializing the prototype memory bank. The performance of some TTA methods, such as ETA and LAME, does not meet the expected standards on Office-Home and other datasets. In fact, they even exhibit inferior performance compared to the source model on certain tasks (\textit{e.g.}, Art, Product, and Real), which highlights the limitations of TTA and the necessity of few-shot target samples. In conclusion, our FS-TTA demonstrates a notable advantage in tasks that closely resemble real-world scenarios and provides a significant boost in performance with minimal additional computational overhead.

Finally, for a more comprehensive comparison with TTA methods, we construct new baselines by combining fine-tuning with representative TTA approaches. Specifically, we select TENT~\cite{wang2020tent}, as a widely adopted and foundational method in test-time adaptation, and TSD~\cite{wang2023feature}, which demonstrates state-of-the-art performance across benchmarks. According to the results in Table~\ref{tab:compare_tta}, our method achieves an average improvement of 4.2\% over Fine-Tuning+TENT and 2.4\% over Fine-Tuning+TSD across the three datasets. These results highlight the superiority of our framework in migrating to the few-shot TTA setting, benefiting from the proposed FDA module and the support-set-based prototype initialization.

% \begin{figure*}[t]
%     \begin{picture}(0,192)
%     \put(0,3){
%     \includegraphics[width=0.97\linewidth]{ab1.pdf}
%     }
%     \end{picture}
%     \caption{Ablation Study on (a) effectiveness analysis about two-stage framework and (b) effectiveness analysis about FDA module.}
%     \label{fig:ab1}
% \end{figure*}

% \begin{figure*}[t]
%     \begin{picture}(0,183)
%     \put(7,3){
%     \includegraphics[width=0.95\linewidth]{ab2.pdf}
%     }
%     \end{picture}
%     \caption{Comprehensive Sensitivity Analysis of (1) Different Entropy Filter Proportion $\alpha$ and K-shot Number on Model Performance.}
%     \label{fig:ab2}
% \end{figure*}

\begin{figure*}[t]
    \centering
    \includegraphics[width=1\linewidth]{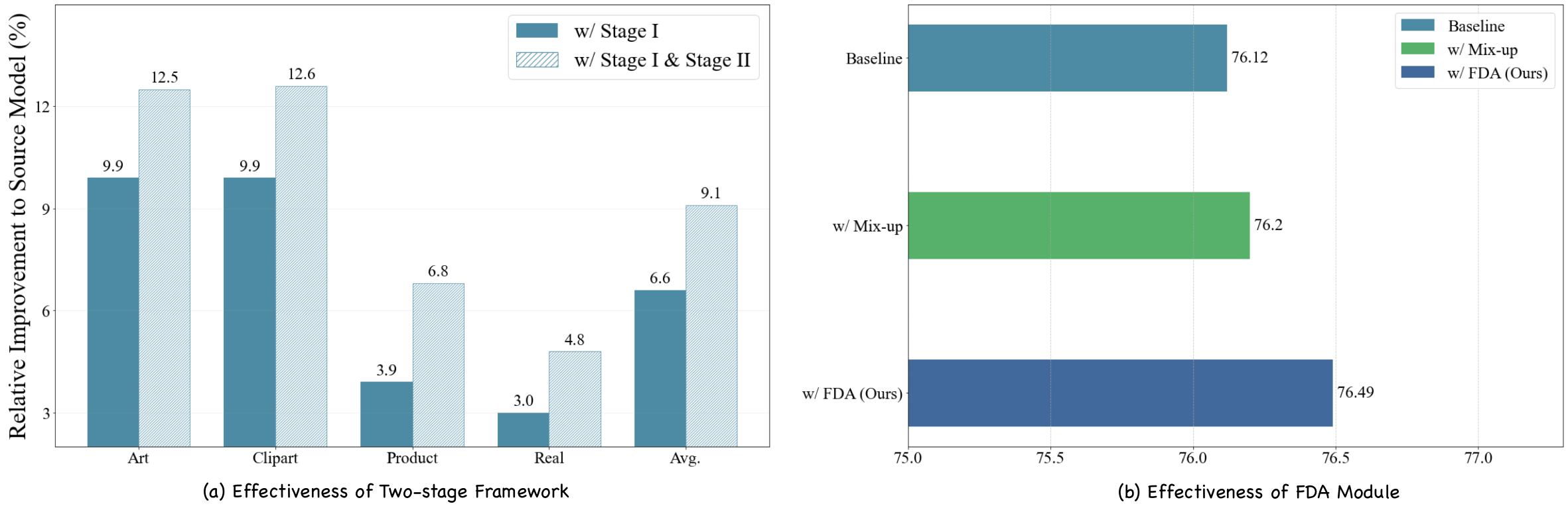}
    \vspace{-0.2in}
    \caption{\textbf{(a)} Effectiveness of the two-stage framework. Stage I already yields notable gains over the source model, while adding Stage II brings further improvements, demonstrating the complementary roles of boundary alignment and distribution refinement.
    \textbf{(b)} Effectiveness of the FDA module. FDA improves few-shot fine-tuning over both the baseline and Mix-up, showing that enhanced feature diversity benefits low-shot adaptation.}
    \label{fig:ab1}
\end{figure*}

\vspace{0.1cm} \noindent \textbf{Comparison with DG/SFDA methods.}
The above experiments mainly focus on TTA, which aims to adapt the model during the test time. A natural question arises: \textit{How about our method compared with domain generalization (DG) or source-free domain adaptation (SFDA) methods? }

To answer this question, we compare our method with several methods in DG and SFDA. The results of Office-Home dataset are shown in Table~\ref{tab: compare_dg}. It can be seen that our method outperforms the state-of-the-art methods in DG, such as SWAD and PCL. In addition, we combine FS-TTA with a SWAD-pretrained model to examine whether our framework can further benefit from a stronger DG-trained initialization.
The resulting SWAD+FS-TTA model achieves 79.8\% accuracy on OfficeHome, indicating that FS-TTA can effectively leverage the flat-minima representations learned by SWAD and yield additional improvements during test-time adaptation.
This demonstrates that FS-TTA is highly compatible with diverse source-model training paradigms, serving as a plug-and-play adaptation module that can be seamlessly integrated with various DG optimizers to further enhance robustness under domain shift.In comparison to advanced SFDA methods, FS-TTA still achieves satisfactory results. It is worth noting that FS-TTA is more flexible in real-world scenarios than SFDA since it adapts the target data in an offline manner, requiring more training loops and resources. The results of DomainNet are shown in Table~\ref{tab: compare_dg}. The overall performance of FS-TTA outperforms the SFDA methods, suggesting that FS-TTA is more adept at handling challenging tasks.

\vspace{0.1cm} \noindent \textbf{Comparison with few-shot transfer learning methods.} In our research, we focus on Few-Shot Test Time Adaptation (FS-TTA), which utilizes a small number of target domain samples to enhance adaptation. To ensure a comprehensive evaluation, we compare our approach with existing few-shot transfer learning methods.  The results on the PACS dataset are presented in Table~\ref{tab:compare_fewshot}. According to the results, FS-TTA consistently outperforms all baseline methods across different domains, achieving the highest average accuracy of 91.4\%, which surpasses the best-performing baseline LCCS (88.8\%). This result highlights the effectiveness of our approach in adapting to domain shifts and improving classification performance in the few-shot setting.

\vspace{-0.1in}
\subsection{Ablation Study}
\label{sec:ablation}
\vspace{0.1cm} \noindent \textbf{Effectiveness of two-stage framework.}
Our proposed method consists of two stages, with the individual contributions of each stage presented in Figure~\ref{fig:ab1}(a). Compared to the baseline source model, Stage I of our approach achieves an average improvement of 6.6\% on the Office-Home. This highlights the effectiveness of our fine-tuning strategy, which employs a mixture of statistics between samples, validating its suitability for the target domain. Our test time adaptation method, which relies on class prototype memory bank guidance during Stage II, adds an extra 2.5\% performance enhancement. As a result, our two-stage framework establishes itself as a robust foundation for the Few-Shot Test Time Adaptation setting, demonstrating its considerable potential in enabling online model adaptation in real-world situations where labeled data is scarce.

\begin{figure*}[t]
    \centering
    \begin{picture}(0,128)
    \put(-270,-2){
    \includegraphics[width=1.02\linewidth]{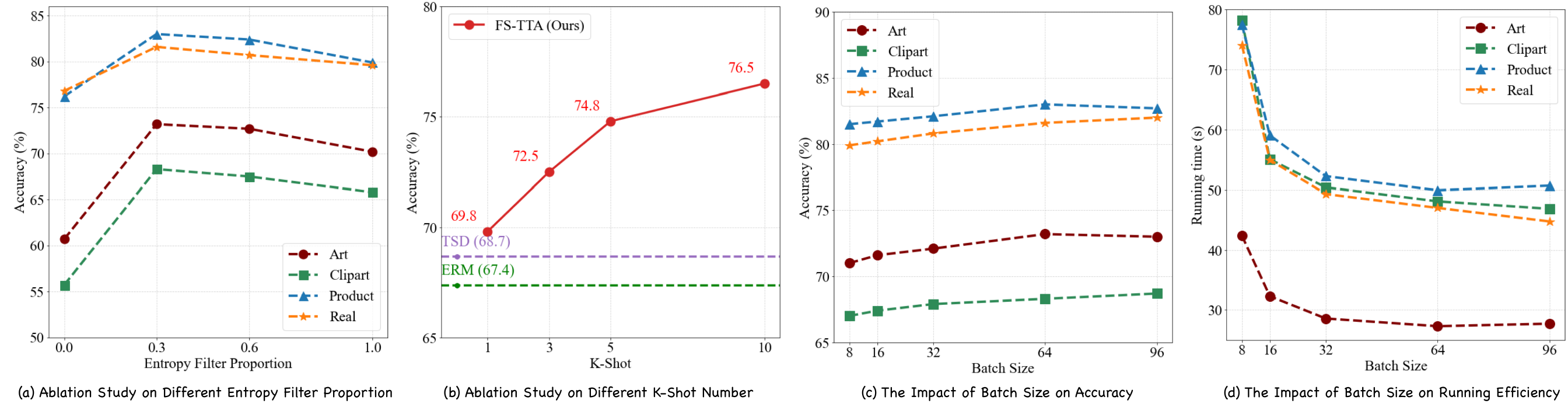}
    }
    \end{picture}
    \vspace{-0.1in}
    \caption{
    \textbf{(a)} Sensitivity of pseudo-label filtering to different entropy-filter proportions $\alpha$. 
    \textbf{(b)} Effect of K-shot supervision strength on adaptation accuracy. 
    \textbf{(c)} Influence of batch size on accuracy during online refinement. 
    \textbf{(d)} Influence of batch size on running efficiency.}
    \label{fig:ab2}
\end{figure*}

% \begin{figure*}[t]
%     \centering
%     \includegraphics[width=0.8\linewidth]{ab2.pdf}
%     \vspace{-0.2cm}
%     \caption{Sensitivity analysis of (1) entropy-filter proportion $\alpha$ and (2) K-shot setting. Analyzing the Impact of Batch Size on Accuracy and Running Efficiency.}
%     \label{fig:ab2}
% \end{figure*}

% \begin{figure*}[t]
% 	\centering
%     \begin{picture}(0,145)
%         \put(-237,10){\includegraphics[width=1\textwidth]{tsne.pdf}}
%     \end{picture}
%     \vspace{-0.3cm}
% 	\caption{ The t-SNE feature visualization of (a) ERM, (b) TENT, (c) LAME, and (d) FS-TTA (Ours).}
% 	\label{fig:tsne}
% \end{figure*}

% \begin{figure*}[t]
%     \begin{picture}(0,183)
%     \put(0,3){
%     \includegraphics[width=0.98\linewidth]{ab3.pdf}
%     }
%     \end{picture}
%     \caption{Analyzing the Impact of Batch Size on Accuracy and Running Efficiency.}
%     \label{fig:ab3}
% \end{figure*}

\begin{figure*}[t]
    \centering
    \includegraphics[width=1.0\textwidth]{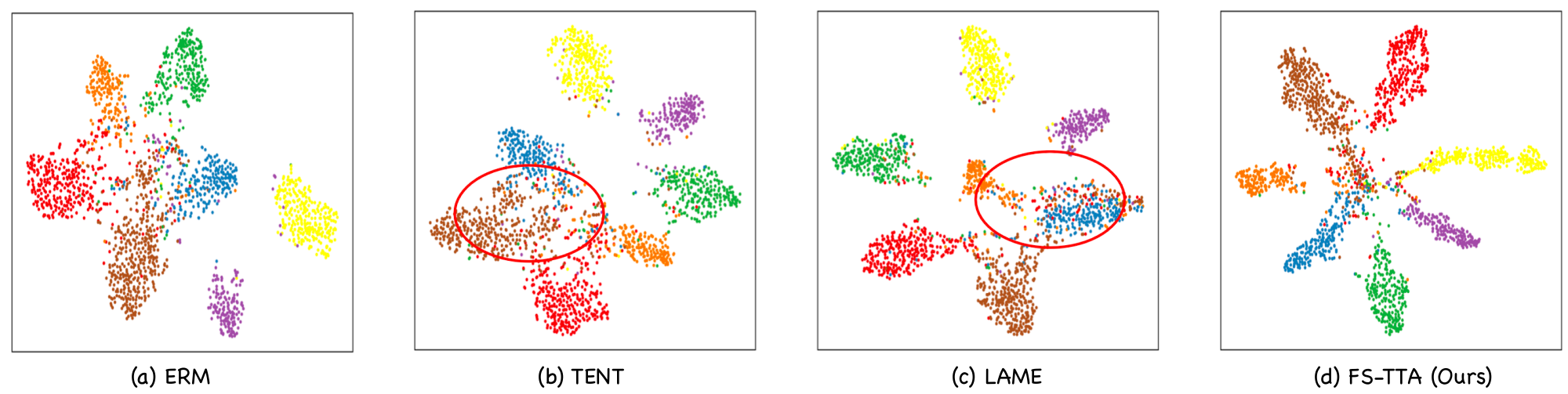}
    \vspace{-0.2in}
    \caption{t-SNE visualization of target-domain features under different adaptation methods. 
    The plots compare the feature distributions produced by (a) ERM, (b) TENT, (c) LAME, 
    and (d) our FS-TTA framework, illustrating how various strategies influence the target-domain 
    representation structure.}
    \label{fig:tsne}
\end{figure*}

% \begin{figure*}[t]
%     \centering
%     \includegraphics[width=0.8\textwidth]{ab3.pdf}
%     \vspace{-0.2cm}
%     \caption{Analyzing the impact of batch size on accuracy and running efficiency.}
%     \label{fig:ab3}
% \end{figure*}
\vspace{0.1cm} \noindent \textbf{Effectiveness of FDA module.}
In our first phase, we introduce the FDA module to tackle overfitting issues through feature augmentation. Here we conduct additional ablation experiments on the FDA module and compare it with Mix-up augmentation, as depicted in Figure~\ref{fig:ab1}(b). The results from the ablation experiments indicate that the FDA module is effective and outperforms mix-up augmentation. The baseline method (without any techniques in the fine-tuning phase) achieves an accuracy of 76.12\%, while incorporating Mix-up leads to a slight improvement, reaching 76.2\% (+0.08\%). However, when our FDA module is introduced, the performance further increases to 76.49\%, yielding a notable improvement of +0.37\% over the baseline. These results highlight the advantage of FDA module in enhancing feature diversity and robustness, surpassing standard augmentation techniques like Mix-up. 

\vspace{0.1cm} \noindent \textbf{Sensitivity to $\alpha$.} The parameter $\alpha$ represents the proportion of each batch that is selected through an entropy filter to update the prototype memory bank and the model. To evaluate the impact of $\alpha$, we conduct an experimental analysis on the Office-Home dataset by assigning $\alpha$ to 0, 0.3, 0.6, and 1, respectively. The results, as shown in Figure~\ref{fig:ab2}(a), demonstrate that $\alpha > 0$ yields performance improvements compared to $\alpha=0$ (the source model), highlighting the effectiveness of our proposed framework. Furthermore, $\alpha=0.3$ and $\alpha=0.6$ perform better than $\alpha=1$ (no filter), indicating the effectiveness of our entropy filter strategy.

 \vspace{0.1cm} \noindent \textbf{Ablation experiments on shot size.} To elucidate the impact of the number of k-shots on our method, we carry out additional ablation experiments within the Office-Home dataset. The findings, illustrated in Figure~\ref{fig:ab2}(b), indicate a significant performance enhancement when the shot size ranges from 1 to 10, demonstrating a rapid performance ascension in this few-shot regime. Remarkably, even minimal shot sizes such as 1-shot and 3-shot exhibit substantial effectiveness. For instance, the 3-shot configuration achieves a 3.8\% performance improvement over the TSD.

\vspace{0.1cm} \noindent \textbf{Efficiency analysis.}
In our main experiments, we opt for a mini-batch size of 64. To examine the variations in performance and computational efficiency with different batch size during test-time adaptation, we conduct a series of analytical experiments. As shown in Figure~\ref{fig:ab2}(c), we observe that accuracy experiences a gradual increase as the batch size incrementally grows, reaching a plateau around a batch size of 64. In contrast, as shown in Figure~\ref{fig:ab2}(d), running time exhibits a decreasing trend as the batch size grows. However, beyond a batch size of 64, the running time appears to stabilize. Consequently, for real-world applications aiming to achieve a trade-off between accuracy and computational efficiency, we suggest a batch size in the vicinity of 64.

\begin{table}[t]
  \centering
  \caption{Comparison of our method with the baseline TSD on both ResNet and ViT-B/16 backbones across the PACS and Office-Home datasets.}
    \resizebox{0.4\textwidth}{!}{
        \begin{tabular}{l|cc}
        \toprule
        Backbones~~~ & ~~PACS~~  & Office-Home \\
        \midrule
        ResNet &84.59 & 67.37 \\
        + TSD~\cite{wang2023feature} &89.41 &68.67 \\
        + Ours & \textbf{91.42} &\textbf{76.49} \\ \midrule
        ViT-B/16 &87.13 &79.06 \\
        + TSD~\cite{wang2023feature} &90.20 &81.80 \\
        + Ours &\textbf{91.89} &\textbf{87.32} \\ \midrule
        \end{tabular}
    }
    \label{tab:vit}
    \vspace{-0.4cm}
\end{table}

 \vspace{0.1cm} \noindent \textbf{Qualitative analysis by t-SNE visualization.}
We present t-SNE
% ~\cite{JMLR:v9:vandermaaten08a} 
visualizations to compare the feature representations of the pre-trained source model (ERM), test time adaptation methods (TENT and LAME), and our proposed method, as illustrated in Figure~\ref{fig:tsne}. The learned features of the pre-trained source model on the target domain are not well-separated due to the significant domain gap, as shown in Figure~\ref{fig:tsne}(a). Additionally, we can observe no considerable feature distribution changes on the target domain after adaptation with TENT and LAME methods, as shown in Figure~\ref{fig:tsne}(b) and Figure~\ref{fig:tsne}(c). In contrast, our method produces more uniform and aligned feature distribution after adapting to the target domain, as shown in Figure~\ref{fig:tsne}(d).

 \vspace{0.1cm} \noindent \textbf{Scalability on Vision Transformer.} We conduct experiments to verify whether our method can be applied to other architectures, such as Vision Transformer (ViT)~\cite{dosovitskiy2020image}. Specifically, we adopt ViT-B/16 as the backbone and compare the baseline TSD with our approach. The results, shown in Table \ref{tab:vit}, demonstrate that our method achieves consistent improvements over TSD. On the PACS dataset, our method improves the accuracy from 90.20\% (TSD) to 91.89\%, while on Office-Home, it further boosts performance from 81.80\% to 87.32\%. These gains highlight that our approach is not limited to convolutional networks but can also optimize transformer-based architectures, making it a versatile solution for various backbone choices.

\section{Conclusion}
\label{sec:conclusion}
In this work, we introduce Few-Shot Test Time Adaptation (FS-TTA), a novel setting that diverges from traditional TTA by leveraging the few-shot support set to improve adaptation to the target. To tackle FS-TTA, we propose an effective framework, which involves employing the few-shot support set to fine-tune the pre-trained source model and maintaining a prototype memory bank to guide the test time adaptation. Results on three cross-domain benchmarks demonstrate the superior performance and reliability of our method. Looking ahead, we aspire to expand FS-TTA beyond current scope by investigating potential real-world tasks, instead of limiting to image recognition.

% \section*{Acknowledgments} 
% The work was supported by the National Natural Science Foundation of China under Grant 62301310.

\bibliographystyle{IEEEtran}
\bibliography{main}

\vfill

\end{document}